# CFL-aware edge design for graph neural solvers of hyperbolic transport

Jinyoung Jeong[1], Joseph B. Choi[2], Xinlun Cheng[2], H. S. Udaykumar[3], Sanghun Choi[1,*], and Stephen S. Baek[1,2,4,*]

[1] School of Mechanical Engineering & Research Institute of Engineering Design & Technology, Kyungpook National University, Daegu, South Korea

[2] School of Data Science, University of Virginia, Charlottesville, VA 22903, USA

[3] Department of Mechanical Engineering, University of Iowa, Iowa City, Iowa 52242, USA

[4] Department of Mechanical and Aerospace Engineering, University of Virginia, Charlottesville, VA 22903, USA

* Corresponding authors: Sanghun Choi (s-choi@knu.ac.kr) and Stephen S. Baek (baek@virginia.edu)

## Abstract

Hyperbolic partial differential equations (PDEs) transport information at finite speed along characteristics, often producing shocks and discontinuities that are challenging to capture numerically. Classical numerical schemes employ reconstruction using spatial stencils to model such sharp gradients in an accurate and numerically stable manner. The reliability of these methods, namely how far a method may advance per time step without loss of accuracy or stability, is governed by a dimensionless ratio, called the Courant–Friedrichs–Lewy (CFL) number, defined as the wave speed times the time step over grid spacing. Recently, graph neural networks (GNNs) have emerged as an appealing alternative to these classical schemes. Their way of gathering and aggregating information from neighboring nodes naturally mirrors the way spatial stencils work in classical numerical schemes, offering a promise of better performance through learned update rules rather than hand-designed ones. However, neighbor-aggregation alone does not tell a GNN how far information must travel and along which edge. As a result, when a GNN model takes a time step larger than what was used in training, shock fronts smear and oscillations grow. In this work, we ask whether making the scale of transport explicit as an edge-local CFL signal can enhance GNNs' ability to model hyperbolic PDEs. Holding the architecture fixed, we vary only the information given to each edge, from a raw edge form to forms that progressively expose the edge-local transport interaction through direction, length, and wave speed. Because all of these are computed from the same ingredients, what changes is not the amount of information but its form. The results show a consistent pattern, in which feature count is not the limiting factor, and appending CFL directly to a raw edge form produces almost no benefit. The CFL conditioning proves most effective when the edge information already exposes the transport interaction, reducing final-step error by 10.4–14.4%, and the improvement widens as the step size grows larger. Notably, this improvement arises not from smearing the solution but from suppressing oscillations near shocks. These results point to a specific design principle that CFL conditioning is most effective when the edge representation already separates the direction of interaction from spatial scale, so that the CFL value has a well-defined quantity to modulate.

## 1. Introduction

Hyperbolic partial differential equations (PDEs) describe a broad class of physical phenomena governed by wave propagation and shock formation, in which information is transported directly rather than diffused [1,2]. A shock front, for example, is a moving discontinuity that moves across the spatial domain at a speed determined by the local state, instead of values simply changing in place. Accurately modeling shocks is, therefore, a problem of accurately modeling transport: how far information moves in one step, in which direction it moves, and which neighboring information should be trusted. As we show in this work, giving a learned solver this transport information is not simply a matter of adding more local features to each edge. Instead, it depends on how that information is organized relative to the transport scale it is meant to represent.

Classical shock-capturing schemes have evolved to handle this transport problem robustly via numerical flux evaluation and local reconstruction. The CFL (Courant–Friedrichs–Lewy) condition limits how far information may travel relative to the local mesh scale in one step [3]. Upwinding and Riemann solvers determine the physically correct direction along which information propagates across interfaces [1,2,4–6]. Essentially non-oscillatory (ENO) and weighted essentially non-oscillatory (WENO) schemes achieve robustness by adaptively selecting or weighting neighboring stencils according to the local measurement of smoothness, thereby maintaining higher-order accuracy while suppressing spurious oscillations near shocks and discontinuities [7,8]. These methods remain standard tools for hyperbolic PDEs and shock-dominated flows.

Despite their success, classical shock-capturing methods require problem-dependent choices of numerical fluxes, reconstruction stencils, and time-step restrictions to achieve stability and accuracy [1,2,7,8]. As geometries and flow regimes become more complex, manually designing these numerical components becomes increasingly difficult. This challenge has motivated the development of neural solvers that retain the local-stencil update structure of classical finite volume methods (FVM) while learning the update rule directly from data [9–16].

Among these approaches, graph neural networks (GNNs) are a particularly natural candidate for neural solvers. A GNN views the computational domain as a graph whose nodes correspond to discrete solution points and whose edges encode local interactions. At each time step, the value at each node is updated by aggregating messages received from its neighboring nodes [17–19]. On a regular, structured grid, the locations of the neighbors used by each point are fixed by the underlying graph topology. On an irregular graph, the number, distance, and direction of the neighbors can vary from node to node. GNNs accommodate this variability in neighbor relations by explicitly representing edge connectivity, whose geometric attributes can encode quantities such as the direction and distance of each local interaction. Consequently, GNNs naturally extend stencil-based local updates to irregular discretizations, making them well-suited for simulations involving complex geometries and nonuniform computational domains [9–12,20].

Because of these advantages, a growing body of work has explored GNN-based neural solvers that predict the evolution of PDE states step-by-step on irregular graphs arising from complex geometries and discontinuities [9–12,20]. These studies show that local message passing can learn temporal evolution while accommodating flexible computational domains. For hyperbolic PDEs, the essential issue is not just aggregating information from neighboring nodes, but representing transport—the direction in which information propagates and the distance over which it travels during one time step. In conventional GNNs, graph edges specify which neighboring nodes may exchange information, but they do not explicitly encode the scale of transport associated with each interaction along the edge. One line of prior work revisited the CFL condition for message-passing neural PDE solvers and showed that larger CFL values require GNNs to aggregate information over larger neighborhoods [21]. However, in this study, we argue that expanding the receptive field alone is insufficient. Equally important, if not more, is what transport information is conveyed by each edge. If the transport scale associated with an edge remains implicit, the GNN must reconstruct it from node values and geometry on its own, and message passing tends to behave in a diffusive manner. Additionally, at large time steps, shock fronts can drift while spurious oscillations develop nearby.

The need for an edge-local CFL is clearest on irregular graphs, where edge lengths and directions vary from edge to edge and shock speed changes with local state and edge direction. Thus, how far information should travel in one step must be recomputed for each edge, not assigned globally. We therefore make this scale explicit as an edge-local CFL value. But simply adding a scalar CFL number is not enough. CFL specifies how far information should move; however, the edge representation must also expose the unit edge direction independently of edge length while retaining edge length as its spatial normalization, so that each CFL value can be associated with the directed edge interaction it quantifies. This is the core idea we test in this paper: rather than merely adding more local information, a GNN can take the same local information and reorganize it into a different form better suited to a transport update, one where CFL can act as a scale that directly modulates a specific already-identified edge interaction. We call this the *compatibility hypothesis*: CFL conditioning is effective when the edge representation explicitly expresses the direction and spatial scale of the local interaction quantified by the CFL, and much less effective otherwise. In other words, a CFL value is not a fact about a single point, but it is the scale of an interaction across an edge.

As an archetypal problem [1,7], we study this phenomenon on the two-dimensional anisotropic scalar Burgers equation on an irregular graph constructed from a periodic irregular point cloud. This setting retains nonlinear transport, state-dependent wave speed, shock formation, edge-dependent geometry, and CFL sensitivity, while avoiding the added complexity of multiple characteristic families [1,2]. It is therefore a controlled testbed for isolating how edge-local CFL and the organization of edge information affect message passing. Our contribution is not to reinvent the CFL number itself, but a controlled ablation of when GNNs can actually use CFL-scale information. More specifically, we show that:

1. Expanding edge features alone leaves the large-time-step accuracy gap open, regardless of how much local geometric detail is added (Section 5.1);
2. The benefit of CFL conditioning depends on how local transport information is represented and whether the edge representation explicitly exposes the local transport interaction (Section 5.2);

3. In representations that expose this interaction, preserving the sign of the CFL number provides little additional benefit beyond its magnitude alone, with the improvement arising from reduced oscillations near shocks rather than increased numerical smoothing (Section 5.3); and
4. The benefit of explicit CFL conditioning persists across message-passing depths and grows with the transport gap, indicating that a larger receptive field cannot replace explicit transport information (Section 5.4).

Our experiments in this paper highlight this distinction. Reorganizing edge information into a more transport-friendly form, on its own, does not sufficiently reduce the transport error that appears at large time steps: even with direction and wave-speed ingredients exposed, the model cannot track shocks robustly without an explicit transport scale. With CFL conditioning, however, the effect depends strongly on the edge form. When edge information is organized closer to a transport update, CFL conditioning reduces the final-step error by 10.4–14.4%, whereas the raw edge form improves only weakly; the representation that explicitly separates edge direction from edge length attains the lowest absolute final-step error. The difference becomes more pronounced as the time step increases and information must travel farther in a single time step. The gain appears not as simple smoothing but as reduced oscillation near shocks, a pattern we later interpret as a property of message passing itself under hyperbolic transport. Figure 1 summarizes the problem setting, the edge-local CFL construction, and the representation progression examined below.

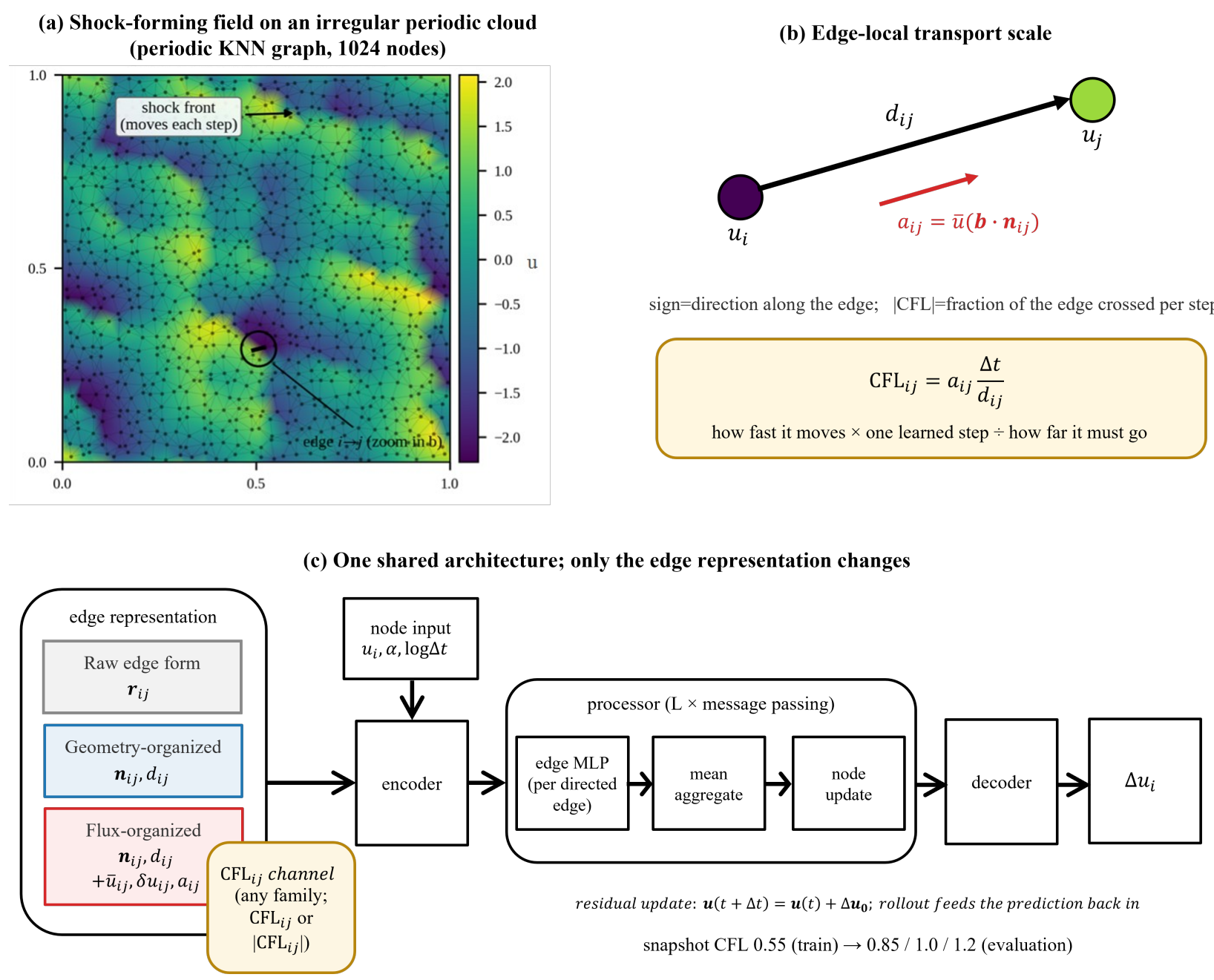

**Figure 1.** Overview of edge-local CFL conditioning on an irregular graph. A shock-forming solution of the two-dimensional anisotropic scalar Burgers equation is sampled on a periodic 1024-node point cloud. For each directed edge, the edge-normal characteristic speed, snapshot interval, and edge length define the signed edge-local CFL. The Raw-edge, Geometry-organized, and Flux-organized forms progressively expose the same local interaction through displacement, explicit direction and length, and local transport quantities, respectively; signed or absolute CFL can then be attached to the corresponding edge as an optional conditioning channel.

# 2. Problem setup and reference data

## 2.1. Governing law

We consider a scalar conservation law on the periodic unit square $\Omega = [0,1]^2$:

$$\partial_t u + \nabla_x \cdot \boldsymbol{f}(u) = 0, \quad \boldsymbol{x} \in \Omega, \quad t \in (0,T]$$
$$u(\boldsymbol{x},0) = u_0(\boldsymbol{x}), \quad \boldsymbol{x} \in \Omega \tag{1}$$

where $u$ is the scalar to be transported. The anisotropic flux vector is defined as:

$$\boldsymbol{f}(u) = \frac{u^2}{2}\boldsymbol{b}, \quad \boldsymbol{b} = \begin{pmatrix} 1 \\ \alpha \end{pmatrix} \tag{2}$$

We set $\alpha = 0.5$ in all experiments.

We use the anisotropic scalar Burgers equation as a minimal shock-forming hyperbolic testbed because it retains nonlinear, state-dependent transport and finite-time shock formation while avoiding the complexity of multiple interacting characteristic families, each with its own characteristic speed [1,2]. Instead, the constant vector $\boldsymbol{b}$ in Eq. (2) specifies a single transport axis and, in smooth regions, a single characteristic velocity $\boldsymbol{f}'(u) = u\boldsymbol{b}$. The initial-condition ensemble, reference solver, snapshot time sampling, and associated CFL quantities are described in the subsequent sections.

## 2.2. Reference solution and snapshot CFL

Reference trajectories are generated on a uniform $128 \times 128$ Cartesian grid ($\Delta x = \Delta y = 1/128$) using a fifth-order WENO finite-difference discretization, Lax–Friedrichs flux splitting, SSP-RK3 time integration [7,8,22], periodic boundary conditions, and a solver CFL of 0.25. A smooth-regime nested-grid verification of the reference implementation is summarized in Appendix A.1. Initial conditions are generated by first drawing a random smooth field: a sum of 10 Fourier modes with integer wavenumbers up to five, random phases, and Gaussian amplitudes, normalized to zero mean and unit standard deviation. The initial field is expressed as

$$\tilde{u}_0(\boldsymbol{x}) = \sum_{m=1}^{10} A_m \sin(2\pi \boldsymbol{k}_m \cdot \boldsymbol{x} + \phi_m), \quad u_0(\boldsymbol{x}) = \frac{\tilde{u}_0(\boldsymbol{x})}{\sigma_G(\tilde{u}_0)}. \tag{3}$$

For each mode $m$, $\boldsymbol{k}_m = \left(\boldsymbol{k}_{(x,m)}, \boldsymbol{k}_{(y,m)}\right)$ is its integer wavevector, $\phi_m$ is its random phase, and $A_m$ is its Gaussian amplitude. In the data-generation procedure, $k_{(x,m)}$ and $k_{(y,m)}$ are sampled uniformly from $\{1, \dots, 5\}$ and $\{0, \dots, 5\}$, respectively, while $\phi_m$ is sampled uniformly from $[0, 2\pi)$ and $A_m$ is sampled from a standard Gaussian distribution. The modes are sampled independently with replacement, and $\sigma_G(\tilde{u}_0)$ denotes the empirical standard deviation over the reference grid.

Training data are then generated by temporally downsampling each reference trajectory. We save solution snapshots at a fixed interval $\Delta t$, which is determined by selecting a nominal snapshot CFL number and deriving $\Delta t$ from

$$\Delta t = \frac{\mathrm{CFL_S}}{\|u_0\|_\infty \left(\frac{1}{\Delta x} + \frac{|\alpha|}{\Delta y}\right)}, \tag{4}$$

where $\|u_0\|_\infty$ denotes the maximum absolute initial-state value on the reference grid, and $\mathrm{CFL_S}$ is the nominal snapshot CFL selected for the trajectory. For training and in-distribution evaluation, $\mathrm{CFL_S} = 0.55$ is used, whereas out-of-distribution evaluation uses $\mathrm{CFL_S} = 0.85$. The larger values $\mathrm{CFL_S} = 1.0$ and $\mathrm{CFL_S} = 1.2$, and therefore larger snapshot intervals, are used for the $\Delta t$ sweep in Section 5.4. These values are selected to examine model generalization with respect to the time step. Within each snapshot interval, the integrator takes adaptive substeps to obtain an accurate solution at the fixed interval $\Delta t$. These substeps satisfy

$$\Delta t_{\mathrm{sub}} \leq \frac{0.25}{\|u\|_\infty \left(\frac{1}{\Delta x} + \frac{|\alpha|}{\Delta y}\right)}, \tag{5}$$

where $\|u\|_\infty$ denotes the maximum absolute state value on the reference grid, evaluated at the current internal substep. The neural model does not observe the intermediate solver substeps; each learned update advances the state from one snapshot to the next. At each snapshot, the reference solution is spatially interpolated from the regular reference grid onto the irregular periodic point clouds using periodic bilinear interpolation, producing the point-cloud representation used for graph learning. Graph construction is presented in the next section, and the complete setup is summarized in Table 1.

**Table 1.** Reference problem, numerical solver, and graph-learning setup.

| Item | Setting |
| --- | --- |
| PDE | Two-dimensional anisotropic scalar Burgers equation |
| Flux | $\boldsymbol{f}(u) = \frac{u^2}{2}\boldsymbol{b}, \quad \boldsymbol{b} = \begin{pmatrix}1\\ \alpha\end{pmatrix}, \quad \alpha = 0.5$ |
| Domain | Periodic unit square |
| Initial conditions | 10 random Fourier modes with integer wavenumbers up to 5; zero mean, unit standard deviation |
| Reference solver | Fifth-order WENO finite-difference discretization + Lax–Friedrichs flux splitting + SSP-RK3 |
| Reference grid | Uniform 128 × 128 Cartesian grid ($\Delta x = \Delta y = 1/128$) |
| Solver CFL | 0.25 |
| Snapshot CFL (training/ID) | 0.55 |

| Item | Setting |
|---|---|
| Snapshot CFL (OOD) | 0.85 (sweep: 1.0, 1.2) |
| Graph | Periodic jittered point cloud |
| Nodes | 1024 |
| Connectivity | Symmetrized periodic KNN; K = 4, 6 |
| Rollout length | 30 autoregressive steps |

### 2.3. Graph construction

The irregular point cloud $V$ is constructed by perturbing a regular lattice of points. First, 1024 nodes ($32 \times 32$) are placed uniformly across the physical domain. Each node is then independently displaced in the $x$- and $y$-directions by an amount sampled uniformly from $[-0.35/32, 0.35/32]$, corresponding to a maximum perturbation of 0.0109375 (0.35 lattice spacings) per coordinate, and the perturbed positions are wrapped periodically into the unit square. Edges are assigned using symmetrized periodic $K$-nearest-neighbor connectivity. Specifically, the $K$ nearest neighbors of each node are selected using periodic minimum-image distances, and each selected connection is symmetrized by adding its reverse connection. Directed edges $E$ are obtained from these symmetrized connections, yielding the graph $G = (V, E)$. For each directed edge $(i, j)$ in $E$, $i$ is the source node, $j$ is the destination node, and the node positions are $\boldsymbol{p}_i$ and $\boldsymbol{p}_j$. The raw displacement is defined as $\Delta\boldsymbol{p}_{(ij)} = \boldsymbol{p}_j - \boldsymbol{p}_i$, and the periodic minimum-image displacement, edge length, and unit edge direction are

$$\boldsymbol{r}_{ij} = \Delta\boldsymbol{p}_{ij} - \mathrm{round}(\Delta\boldsymbol{p}_{ij}), \quad d_{ij} = \|\boldsymbol{r}_{ij}\|_2, \quad \boldsymbol{n}_{ij} = \frac{\boldsymbol{r}_{ij}}{d_{ij}}. \tag{6}$$

Here, round is applied componentwise, yielding a minimum-image displacement on the unit two-dimensional torus and thereby enforcing periodic boundary conditions on the displacement, edge length, and unit edge direction for each directed edge $(i, j)$.

Together, the flux law in Section 2.1, the snapshot interval defined in Section 2.2, and the directed-edge geometry defined here determine the edge-local CFL introduced in Section 3.1.

## 3. Method

### 3.1. Edge-local CFL conditioning and controls

Some of the GNN models compared in this study receive local transport information as part of their input in the form of an edge-local CFL number. We therefore define it here. For each directed edge $(i, j)$ in $E$, define the edge-averaged state and state jump as

$$\bar{u}_{ij} = \frac{u_i + u_j}{2}, \quad \delta u_{ij} = u_j - u_i. \tag{7}$$

From the flux definition in Eq. (2), the directional characteristic speed along this directed edge is

$$a_{ij} = \boldsymbol{f}'(\bar{u}_{ij}) \cdot \boldsymbol{n}_{ij} = \bar{u}_{ij}(\boldsymbol{b} \cdot \boldsymbol{n}_{ij}) = \bar{u}_{ij}(n_{ij,x} + \alpha n_{ij,y}). \tag{8}$$

In addition to this signed average speed, the flux-organized representation uses the local maximum endpoint wave-speed magnitude $s_{ij} = \max(|u_i(\boldsymbol{b} \cdot \boldsymbol{n}_{ij})|, |u_j(\boldsymbol{b} \cdot \boldsymbol{n}_{ij})|)$, a Rusanov-type local signal-speed estimate [1,2,23].

The edge-local signed CFL $\mathrm{CFL}_{ij}$ is then defined as

$$\mathrm{CFL}_{ij} = \frac{a_{ij}\Delta t}{d_{ij}}. \tag{9}$$

This value is supplied as part of the input to models with "+ signed CFL" in their names.

Under edge reversal, the signed CFL is antisymmetric, $\mathrm{CFL}_{ji} = -\mathrm{CFL}_{ij}$, whereas its magnitude is invariant, $|\mathrm{CFL}_{ji}| = |\mathrm{CFL}_{ij}|$. Thus, signed CFL is orientation-sensitive, while absolute CFL retains only the edge-local CFL magnitude.

To isolate the contribution of the sign of the edge-local CFL from its magnitude, we compare models that receive the magnitude of the signed-CFL feature with its sign removed—the edge-local absolute CFL $\mathrm{CFL}_{ij}^{\mathrm{abs}}$. To test edge correspondence, we also use controls in which signed-CFL or absolute-CFL values are randomly permuted across directed edges within a graph snapshot. The signed shuffled control is $\mathrm{CFL}_{ij}^{\mathrm{shuf}}$. These quantities are defined as

$$\mathrm{CFL}_{ij}^{\mathrm{abs}} = |\mathrm{CFL}_{ij}|, \quad \mathrm{CFL}_{ij}^{\mathrm{shuf}} = \mathrm{CFL}_{\pi(i,j)}, \quad \mathrm{CFL}_{ij}^{\mathrm{shuf,abs}} = |\mathrm{CFL}|_{\pi(i,j)}. \tag{10}$$

where $|x|$ denotes the absolute value of $x$, and $\pi$ denotes a fixed random permutation of the directed-edge set $E$ within each graph snapshot. Both shuffled controls preserve the marginal distribution of their respective CFL channel while breaking its edge assignment.

An additional random-noise control appends one uninformative scalar per edge: an independent standard-normal draw during training and a deterministic pseudorandom value in $[-1,1]$ during evaluation. This control matches the dimensionality of a CFL channel but carries no transport information.

## 3.2. Shared message-passing architecture

Within each fixed message-passing depth, all feature variants use the same message-passing architecture on the same directed graph; only the edge representation supplied to the model changes, so even the raw-edge family passes messages along the same directed edges. The node input is fixed as

$$\boldsymbol{x}_i = [u_i, \alpha, \log \Delta t]^{\mathrm{T}}. \quad (11)$$

Here $\alpha$ is the directional flux weight in Eq. (2) and $\Delta t$ the snapshot interval. The $\alpha$ input is included so that the same node parametrization extends to varying flux anisotropy; in the present experiments $\alpha$ is fixed at 0.5, so it acts as a constant input rather than an informative feature.

At each message-passing step, directed edge states are updated from source-node, target-node, and edge features [9–11,17,19,24]. Incoming edge states are mean-aggregated, and each node state is updated from its previous state and the aggregated message. The decoder predicts a scalar increment at each node; collectively, these increments define the state-update vector in Eq. (12). Thus the model is a learned explicit one-step map evaluated by one forward pass. The update is a nonlinear map, but it is not produced by solving an implicit (linear or nonlinear) algebraic system; it does not enforce conservation by construction [25]—edge messages are not constrained to an antisymmetric flux difference, so in general $m_{ij} \neq -m_{ji}$—and has no formal stability guarantee.

$$\hat{\boldsymbol{u}}(t + \Delta t) = \boldsymbol{u}(t) + \Delta \boldsymbol{u}_\theta(\boldsymbol{u}(t), G, \Delta t). \quad (12)$$

The message-passing depth L, defined as the number of message-passing steps, sets the receptive field of the update: after L message-passing steps, a node can depend on information from its L-hop neighborhood. This is stencil-like support, but the learned update is nonlinear, state-dependent, and not a fixed finite-difference operator.

All experiments use a hidden dimension of 64. The encoder, edge-update, node-update, and decoder networks use MLPs with two hidden SiLU layers and a final linear layer. The encoder maps node inputs and edge inputs to 64-dimensional latent vectors. Layer normalization is applied to the encoder and processor MLP outputs. At each message-passing step, the edge MLP receives the concatenation of source-node, target-node, and current edge latent vectors; the node MLP receives the concatenation of the previous node latent and the mean-aggregated incoming edge messages. Both edge and node latent states are updated residually. The decoder concatenates the final node latent with the original node input and maps this skip-augmented representation to a scalar increment $\Delta u$. Mean aggregation is used rather than sum aggregation so that KNN4 and KNN6 models differ in stencil content rather than aggregate scale.

## 3.3. Edge-representation progression and variants

Table 2 organizes the edge representations along a progression from raw inputs to increasingly explicit transport organization. We refer to the three families as *Raw-edge*, *Geometry-organized*, and *Flux-organized* throughout, abbreviated "Raw," "Geometry," and "Flux" in compound model labels (e.g., "Geometry + |CFL|"). The raw-edge representation gives the model the states at the source and destination nodes, the minimum-image displacement in its raw vector form, and the corresponding edge distance, and asks whether message passing can compose local transport from these inputs alone. Attaching signed CFL or |CFL| to this representation tests whether an explicit transport scale can be used before the corresponding interaction coordinates are made explicit. Although each endpoint state is already encoded on its node, all three families also provide the same state information to the edge encoder. The Raw-edge and Geometry-organized families use the source and target states directly, whereas the Flux-organized family uses their equivalent mean and jump; this keeps endpoint-state access fixed and isolates how the same local information is organized on each edge.

The geometry-organized representation makes the edge interaction explicit without adding information. It replaces the raw displacement components with the unit edge direction while retaining the edge length of Section 2.3. Direction specifies the angular orientation of the interaction, whereas length supplies its spatial scale and normalization. This does not change the graph topology or align edges with the flow: the same minimum-image displacement determines both direction and length through Eq. (6). Thus, this transition tests organization alone; the same geometric content is factorized into the interaction coordinates used by the CFL construction.

The flux-organized representation layers local transport quantities on top of that geometry. It uses the edge-averaged state, state jump, unit edge direction, edge length, signed edge-normal characteristic speed, and local maximum endpoint wave-speed magnitude defined in Sections 2.3 and 3.1: the ingredients a numerical flux would use. Without the CFL feature, however, the model must still compose them into a transport scale. The CFL variants, attached within every family, provide that scale explicitly: the edge-local signed CFL defined in Eq. (9) quantifies how far information should move along the corresponding directed edge in one learned update. Because it is computed from that edge's direction, length, local state, and snapshot interval, it is not an interchangeable extra number. The signed- and absolute-CFL correspondence controls and the noise control test these roles by detaching values from their edges and by adding an uninformative channel, respectively (Section 3.1).

**Table 2.** Main CFL-centered ablation matrix. Within each fixed message-passing depth, all variants share the same node input, architecture, optimizer, and rollout protocol. Models are grouped by edge-representation family and then by CFL variant; the CFL variants are labeled by the signed edge-local CFL and its magnitude. “No CFL” means that no edge-local CFL channel is supplied; every variant still advances over the snapshot interval set by the snapshot CFL. Here, $s_{ij}$ denotes the local maximum endpoint wave-speed magnitude defined in Section 3.1.

| Family | Base edge input | CFL variants |
|---|---|---|
| Raw-edge | $u_i, u_j, \mathbf{r}_{ij}, d_{ij}$ | No CFL; $+\,\mathrm{CFL}_{ij}$; $+\,\lvert\mathrm{CFL}_{ij}\rvert$ |
| Geometry-organized | $u_i, u_j, \mathbf{n}_{ij}, d_{ij}$ | No CFL; $+\,\mathrm{CFL}_{ij}$; $+\,\lvert\mathrm{CFL}_{ij}\rvert$ |
| Flux-organized | $\bar{u}_{ij}, \delta u_{ij}, \mathbf{n}_{ij}, d_{ij}, a_{ij}, s_{ij}$ | No CFL; $+\,\mathrm{CFL}_{ij}$; $+\,\lvert\mathrm{CFL}_{ij}\rvert$ |

Comparisons follow the progression: no-CFL families first, then each CFL variant against its own no-CFL parent, with signed- and absolute-CFL shuffling and noise serving only as controls. This keeps cross-family ranking separate from within-family effect comparisons. The experiment therefore tests compatibility between an edge representation and the local time-scaled transport quantity, rather than whether a larger feature vector is always better. No deployable variant adds information: every edge input, including the CFL, is a deterministic function of the endpoint states, positions, flux parameters, and snapshot interval already available to the raw-edge family; only the shuffled and noise controls break this tie. The progression varies organization, not content. We call the geometry-organized and flux-organized families transport-organized because both explicitly expose edge direction and length; the flux-organized family additionally exposes local state and characteristic-speed structure.

# 4. Numerical experiments and evaluation

## 4.1. Training protocol and main experiment matrix

The experiment matrix is designed for paired, interpretable comparisons: within each matched graph realization, connectivity, and message-passing depth, the data, reference solver, architecture, optimizer, and rollout protocol are held fixed, and only the edge representation and edge-local CFL variant change.

Training uses the AdamW optimizer [26] with a constant learning rate of $10^{-3}$, weight decay $10^{-5}$, and gradient-norm clipping at 1.0 [27], for 120 epochs. Each training sample is a single-step pair $\big(u(t), u(t+\Delta t)\big)$ on one graph, and each gradient step uses one such sample. Inputs are not additionally normalized; the initial conditions are zero-mean and unit-variance by construction (Section 2.2). Model selection uses the four validation trajectories: after every epoch we record the validation next-step MSE plus an 8-step rollout MSE, and the checkpoint with the lowest sum is used for all evaluations. Evaluation uses 30-step autoregressive rollout [28,29] without teacher forcing. Every variant is trained independently for each graph realization, connectivity, and message-passing depth. The full matrix is summarized in Table 3, with L4 used for the primary analysis and L2 for the robustness analysis.

**Table 3.** Main experiment matrix: snapshot CFL regimes, graph settings, and message-passing depths.

| Axis | Values |
|---|---|
| Snapshot CFL | 0.55 (training/ID); 0.85 (primary OOD); 1.0 and 1.2 (transport-gap sweep) |
| Graph realizations | 3 (seeds 42, 43, 44) |
| Connectivity | KNN4, KNN6 |
| Message-passing depth | L4 (primary); L2 (robustness); L1 and L6 (auxiliary sweep) |

| Axis | Values |
|---|---|
| Model variants | L2: 9 main variants (Table 2) + 3 signed-CFL shuffling controls + 1 random-noise control; L4: the same 13 variants + 3 CFL-magnitude shuffling controls; L1/L6: 9 main variants only |
| Trajectories per setting | 28 training; 4 validation; 8 ID; 8 primary OOD |
| Rollout evaluation | 64 snapshots; 30 autoregressive steps |

Case composition. For each graph realization seed s in {42, 43, 44} and connectivity K in {4, 6}, the data-generation script creates one irregular periodic point cloud with 1024 nodes and WENO reference trajectories: 28 training, 4 validation, 8 in-distribution probe trajectories (nominal snapshot CFL = 0.55), and 8 larger-step OOD probe trajectories (nominal snapshot CFL = 0.85). Each trajectory contains 64 snapshots, and rollout evaluation uses 30 autoregressive steps.

The main statistical unit for the primary claim is the seed–KNN setting at message-passing depth L4. For each model, the L4 matrix contains 3 graph realizations × 2 KNN connectivities = 6 evaluated settings. Main-text tables report the focused comparisons defined in Section 5.

Section 5.4 reports an auxiliary sweep over message-passing depths L1, L2, L4, and L6 to test whether increasing the message-passing receptive field can replace explicit edge-local CFL conditioning; these runs are not part of the primary matrix. KNN6 L4 is used for representative visualizations because it pairs the densest connectivity with the primary depth. The graph realizations serve as matched experimental settings for the robustness of the CFL-conditioning effect.

The six primary reference datasets (three seeds × two connectivities) contain 288 training, validation, ID, and primary-OOD trajectories. For the L2/L4 matrix, each of the 13 shared model variants is trained independently for every (seed, KNN) pair; three additional CFL-magnitude correspondence controls with shuffled edge assignments are trained at L4 (174 trained models in total, 96 of them in the primary L4 analysis).

Each model is evaluated on the 8 larger-step OOD trajectories of its setting, i.e., 48 OOD rollouts per model at L4.

## 4.2. Accuracy metrics

The evaluation is organized around global rollout accuracy and shock-focused diagnostics. Initial conditions are normalized to zero mean and unit variance (Section 2.2), so reported MSE values can be read in those units: a final-step MSE of 0.031 corresponds to an RMSE of about 0.18. The fields are not re-normalized at each time step; MSE is computed directly in the field scale set by the normalized initial condition.

At rollout step k, the stepwise mean-squared error is

$$\mathrm{MSE}_k = \frac{1}{N}\sum_{i=1}^{N}\left(\widehat{u_i^k} - u_i^k\right)^2. \tag{13}$$

The final-step MSE and average rollout MSE are

$$\mathrm{MSE}_{\mathrm{final}} = \mathrm{MSE}_T, \qquad \mathrm{MSE}_{\mathrm{avg}} = \frac{1}{T}\sum_{k=1}^{T}\mathrm{MSE}_k\,, \qquad T = 30. \tag{14}$$

## 4.3. Shock and oscillation diagnostics

MSE alone cannot tell whether a model handles shocks by moving them correctly, smoothing them away, or leaving oscillations behind [7,8,30,31]. To separate these behaviors, we track, at the final rollout step, the shock-region error on steep-gradient reference nodes and overshoot/undershoot beyond the reference extrema.

Let $G = (V, E)$ be the directed graph with $N = |V|$ nodes, let $\boldsymbol{u}$ and $\widehat{\boldsymbol{u}}$ denote the reference and predicted fields at a given rollout step, and let $d_{ij}$ be the periodic edge length of Eq. (6). For each node i, let $\mathcal{N}(i)$ denote its neighboring-node set in the symmetrized graph. The local steepness at a node is the length-normalized mean incident jump [32,33]:

$$g_i = \frac{1}{|\mathcal{N}(i)|}\sum_{j\in\mathcal{N}(i)}\frac{|u_j - u_i|}{d_{ij}}, \qquad i \in V. \tag{15}$$

Collecting the local steepness values into the nodal vector $g = (g_i)_{i\in V}$, the shock set uses the 90th percentile to select the steepest reference nodes at the final step, and the shock-region error is the MSE restricted to that set:

$$S = \{i \in V: g_i \geq P_{90}(\boldsymbol{g})\}, \qquad \mathrm{MSE}_{\text{shock}} = \frac{1}{|S|}\sum_{i \in S}(\widehat{u_i} - u_i)^2. \tag{16}$$

Overshoot and undershoot measure the largest excursion beyond the reference frame extrema; the positive-part operator in Eq. (17) clips negative excursions to zero:

$$O = \max_{i \in V}\left(\widehat{u_i} - \max_{\ell \in V} u_\ell\right)_+, \qquad U = \max_{i \in V}\left(\min_{\ell \in V} u_\ell - \widehat{u_i}\right)_+. \tag{17}$$

The shock-focused diagnostics are evaluated at the final rollout step. The shock-region mask and reference extrema are computed from the WENO reference field, so every model is compared against the same reference structure; overshoot and undershoot are reported directly.

Within the reference-defined shock mask, we compute for each node the mean |CFL| over all directed edges entering that node, using the final reference field. We rank the nodes by this value. The upper-ranked half is defined as the higher-transport group, and the lower-ranked half as the lower-transport group. We report the higher-transport subset's share of the no-CFL final-step squared error and the within-family relative MSE reduction from + |CFL| separately for the two subsets, summarized over the six matched seed–KNN settings. In a separate evaluation-time intervention, |CFL| is recomputed at each rollout step and replaced by its within-graph mean on either the highest-|CFL| edge decile or an equal-sized random edge subset, without retraining.

**Table 4.** Primary metrics and reporting direction.

| Group | Metric | Report | Better |
|---|---|---|---|
| Accuracy | Final-step MSE | value at step T | lower |
| Accuracy | Average rollout MSE | mean over 30 steps | lower |
| Accuracy | Shock-region MSE | masked steep-gradient error | lower |
| Oscillation | Overshoot/undershoot | extreme violation | lower |

# 5. Results: when edge-local CFL conditioning becomes useful

The results are organized around four main research questions. First, do richer edge ingredients improve larger-step rollout without CFL conditioning? Second, does the benefit of CFL conditioning depend on how the same local information is organized and on whether each CFL value remains attached to its own edge? Third, once edge direction is explicit, does CFL sign add value beyond its magnitude, and does the gain reduce shock oscillations rather than merely smooth the solution? Finally, does the gain grow with the snapshot transport gap and persist across message-passing depths? The comparisons below use the six core models—the no-CFL and + |CFL| variants of the raw-edge, geometry-organized, and flux-organized families—except for the targeted signed-CFL and correspondence controls.

All errors are per-node MSE in normalized field units, so a final-step MSE of 0.031 corresponds to RMSE ≈ 0.18. The in-distribution probe at the training snapshot CFL of 0.55 does not sharply separate the models: final-step errors remain between 0.006 and 0.009. We therefore focus on the larger-step OOD regime at snapshot CFL 0.85, extended to 1.0 and 1.2 in Section 5.4.

## 5.1. Richer edge features alone do not close the larger-step gap

The first comparison here asks whether progressively exposing edge geometry and flux-style transport ingredients is sufficient before an explicit CFL scale is supplied. Table 5 shows that it is not: the raw-edge and geometry-organized families have nearly the same final MSE (0.0311 and 0.0313, respectively), while the flux-organized family is no better (0.0349) despite carrying more transport ingredients. Richer edge ingredients alone therefore do not close the larger-step accuracy gap. In the next section, we test whether explicitly supplying the edge-local transport scale changes this outcome.

**Table 5.** No-CFL feature-family baseline for larger-step OOD WENO rollouts (L4, KNN4/6, seeds 42–44). The three families differ only in how the available edge information is organized.

| Feature family | Final MSE | Avg MSE |
|---|---|---|
| Raw-edge | **0.0311±0.0072** | 0.0134±0.0020 |
| Geometry-organized | 0.0313±0.0068 | 0.0138±0.0023 |
| Flux-organized | 0.0349±0.0076 | 0.0152±0.0025 |

## 5.2. CFL gain depends on representation organization and edge correspondence

Table 6 and Figures 2–3 compare each no-CFL parent with its + |CFL| counterpart. The raw-edge family gains only 2.8%, with improvements in three of the six matched settings. Geometry + |CFL| reduces final-step MSE by 10.4% and attains the lowest absolute error among all evaluated variants (0.0281). Flux + |CFL| produces the largest within-family response, improving by 14.4% in all six settings. Its absolute error nevertheless remains higher than that of Geometry + |CFL|. The flux result therefore measures conditioning sensitivity rather than superior final accuracy. Comparing the within-family |CFL| gains across the same six matched settings shows that the gains are larger for the Geometry- and Flux-organized families than for the Raw-edge family in every setting (Appendix A). Together, these comparisons show that the CFL benefit depends on representation organization rather than feature count alone.

**Table 6.** Core CFL-magnitude comparison for larger-step OOD WENO rollouts (L4, KNN4/6, seeds 42–44). Each family is read as its no-CFL parent followed by its CFL-magnitude variant. Δ final MSE is relative to the no-CFL parent (negative favors conditioning), and Wins counts the matched (seed, KNN) settings in which conditioning lowers MSE.

| Family | CFL variant | Final MSE | Avg MSE | ΔMSE vs parent | Wins | Rel gain (%) |
|---|---|---|---|---|---|---|
| Raw-edge | No CFL | 0.0311±0.0072 | 0.0134±0.0020 | – | – | – |
| | + $\lvert \mathrm{CFL}_{ij} \rvert$ | **0.0304±0.0080** | 0.0127±0.0029 | -0.00070 | 3/6 | +2.8% |
| Geometry-organized | No CFL | 0.0313±0.0068 | 0.0138±0.0023 | – | – | – |
| | + $\lvert \mathrm{CFL}_{ij} \rvert$ | **0.0281±0.0071** | 0.0118±0.0019 | -0.00324 | 5/6 | +10.4% |
| Flux-organized | No CFL | 0.0349±0.0076 | 0.0152±0.0025 | – | – | – |
| | + $\lvert \mathrm{CFL}_{ij} \rvert$ | **0.0296±0.0054** | 0.0117±0.0017 | -0.00523 | 6/6 | +14.4% |

Because pooled standard deviations are comparable to the mean differences, we read the primary effects from paired comparisons across the six matched (seed, KNN) settings (Table 6, Figure 3). Win counts summarize consistency across these settings.

These representation-dependent gains motivate the composition-burden interpretation developed in Section 6.1. Figure 2 shows the corresponding rollout-error evolution.

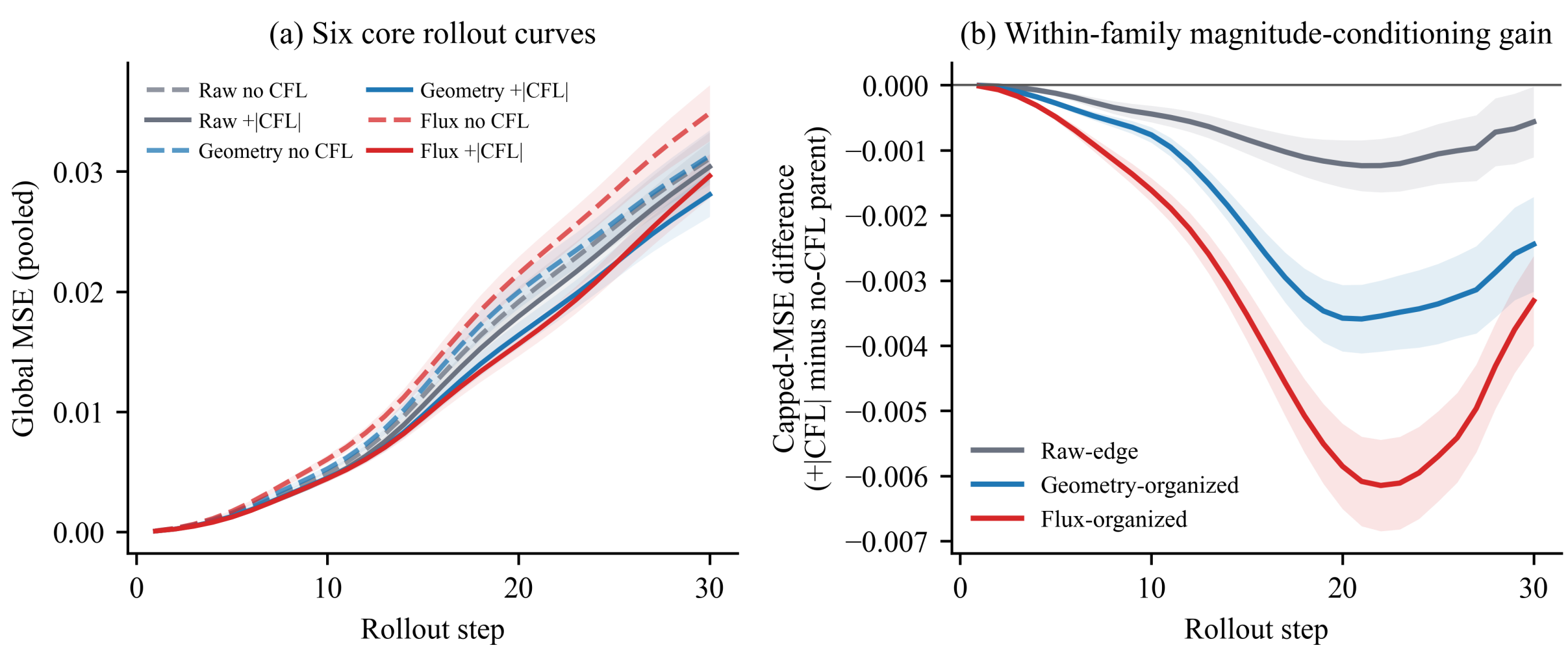

**Figure 2.** Larger-step OOD rollout for the six core models, pooled over the six seed-KNN settings at L4 (shaded: SEM over 48 OOD trajectories). (a) Global rollout MSE over 30 autoregressive steps for the no-CFL and + |CFL| variants of the raw-edge, geometry-organized, and flux-organized families. (b) Within-family capped-MSE difference between each + |CFL| model and its no-CFL parent; curves below zero favor |CFL| conditioning. In panel (b), each trajectory's stepwise MSE is capped at 0.05 before differencing so that a small number of large-error rollouts do not dominate the visualization; all tabulated metrics use uncapped MSE.

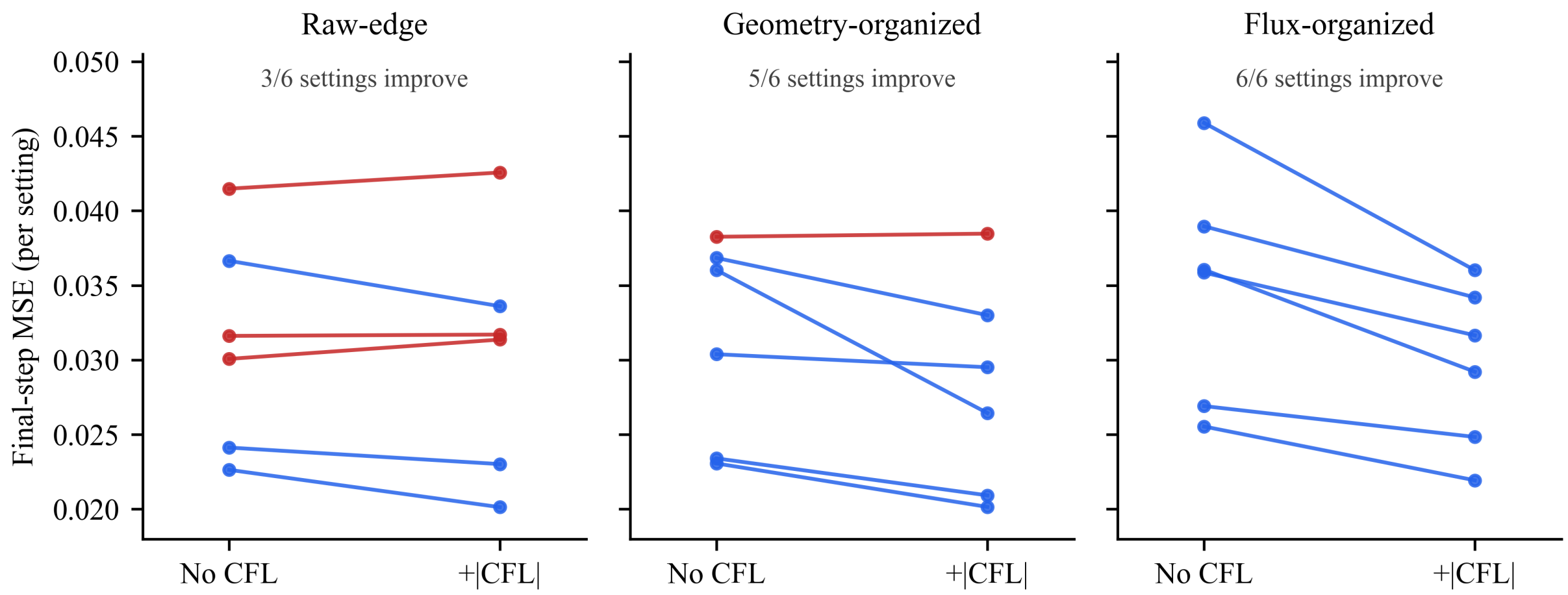

**Figure 3.** Per-setting paired comparison of final-step MSE between each no-CFL parent and its + |CFL| variant (L4, larger-step OOD, snapshot CFL = 0.85). Each line is one matched (seed, KNN) setting; blue indicates improvement and red indicates no improvement. Win counts correspond to Table 6.

Table 6 establishes the representation-dependent benefit of |CFL| conditioning. We next test whether this benefit requires each conditioning value to remain attached to the directed edge on which it was computed, and then repeat the same correspondence control for signed CFL. Each shuffled control preserves the corresponding channel's marginal value distribution while randomizing its assignment across edges (Table 7).

**Table 7.** Edge-local correspondence controls for larger-step OOD WENO rollouts (L4, KNN4/6, seeds 42–44): (a) matched CFL magnitude versus shuffled CFL magnitude and (b) matched signed CFL versus shuffled signed CFL. In both panels, ΔMSE is matched minus shuffled; negative values favor correct edge correspondence.

| **(a) CFL-magnitude correspondence** | | | |
|---|---|---|---|
| **Family** | **ΔMSE (matched - shuffled)** | **Wins matched** | **Rel matched gain (%)** |
| Raw-edge | -0.00170 | 4/6 | +5.9% |
| Geometry-organized | -0.00284 | 4/6 | +8.5% |
| Flux-organized | -0.00771 | 6/6 | **+19.8%** |
| **(b) Signed-CFL correspondence** | | | |
| **Family** | **ΔMSE (matched - shuffled)** | **Wins matched** | **Rel matched gain (%)** |
| Raw-edge | -0.00099 | 3/6 | +1.2% |
| Geometry-organized | -0.00087 | 5/6 | +2.6% |
| Flux-organized | -0.00354 | 5/6 | **+10.6%** |

Correct edge correspondence improves the |CFL| channel by mean per-setting gains of 5.9%, 8.5%, and 19.8% for the raw-edge, geometry-organized, and flux-organized families. Matched variants win 4/6, 4/6, and 6/6 settings, respectively (Table 7(a)). Repeating the control for signed CFL gives gains of 1.2%, 2.6%, and 10.6%, with wins of 3/6, 5/6, and 5/6 (Table 7(b)). Shuffling weakens both encodings on average, with the strongest and most consistent dependence in the flux-organized family.

## 5.3. CFL conditioning reduces shock oscillations, while the sign adds little

With edge correspondence preserved, this subsection asks two targeted questions: whether retaining the CFL sign adds value beyond magnitude when edge direction is already explicit, and whether the resulting improvement controls shock oscillations without relying on excessive smoothing.

Both variants retain explicit edge direction, so this comparison isolates the additional value of the CFL sign beyond its magnitude. In the geometry- and flux-organized families, |CFL| improves over signed CFL by 5.4% and 7.2% on average, respectively (Table 8). The sign of the projected characteristic speed therefore adds little in this single-characteristic-family benchmark. We restrict this targeted comparison to the transport-organized families because they expose edge direction explicitly.

**Table 8.** Targeted signed-CFL versus CFL-magnitude comparison for the transport-organized families (larger-step OOD; L4, KNN4/6, seeds 42–44). ΔMSE is CFL magnitude minus signed CFL; negative values favor the magnitude variant.

| Family | ΔMSE (magnitude - signed) | Wins magnitude | Rel magnitude gain (%) |
|---|---|---|---|
| Geometry-organized | -0.00150 | 5/6 | **+5.4%** |
| Flux-organized | -0.00269 | 4/6 | **+7.2%** |

Shock-region MSE and one-sided excursion diagnostics test whether the same gain occurs at steep fronts. Table 9 reports the six core models, Figure 4 shows the representative final field and local error, and Figure 5 summarizes the corresponding one-sided excursion diagnostics.

**Table 9.** Shock-region final-step MSE for the six core models (larger-step OOD; L4, KNN4/6, seeds 42–44). The shock mask selects nodes whose graph-gradient exceeds the 90th percentile of the reference field. Values are pooled mean ± std over the six matched settings.

| Model | Shock MSE | Wins vs parent |
|---|---|---|
| Raw-edge | 0.2457±0.0559 | – |
| Raw + $\left\|\mathrm{CFL}_{ij}\right\|$ | **0.2394±0.0552** | 4/6 |
| Geometry-organized | 0.2509±0.0525 | – |
| Geometry + $\left\|\mathrm{CFL}_{ij}\right\|$ | **0.2310±0.0549** | 4/6 |
| Flux-organized | 0.2722±0.0549 | – |
| Flux + $\left\|\mathrm{CFL}_{ij}\right\|$ | **0.2402±0.0401** | 6/6 |

Low global rollout error alone does not distinguish accurate shock transport from smoothing or oscillation. Every + |CFL| variant lowers pooled shock-region MSE relative to its no-CFL parent, most consistently for the flux-organized family (6/6 settings; Table 9); Geometry + |CFL| nevertheless attains the lowest pooled shock-region MSE (0.2310). The transport-organized + |CFL| models also reduce both overshoot and undershoot (Figures 4–5).

Overshoot and undershoot measure excursions beyond the reference extrema. Their reduction directly indicates weaker oscillation, while the representative field and cut in Figure 4 show that the shock is not broadly smeared.

Within the reference-defined shock region, the higher-transport half of the nodes accounts for 73.8% and 74.0% of the no-CFL final-step squared error in the geometry-organized and flux-organized families, respectively. The relative error reduction from |CFL| conditioning is also larger in this subset: 10.7% versus 0.4% and 13.7% versus 4.3%, respectively, with the higher-transport subset receiving the larger reduction in all six matched settings (Table 10).

**Table 10.** Transport-stratified shock-region analysis for the geometry-organized and flux-organized families (larger-step OOD; L4, KNN4/6, seeds 42–44). Within the final-step reference shock mask, nodes are ranked by mean incoming-edge |CFL|. The upper-ranked half is defined as the higher-transport group, and the lower-ranked half as the lower-transport group. Error share is measured for the no-CFL parent; relative gain compares + |CFL| with that parent. Values are mean ± std over the six matched settings.

| Family | Higher-transport error share (%) | Relative gain, higher / lower (%) | Higher gain > lower |
|---|---|---|---|
| Geometry-organized | 73.8 ± 3.9 | 10.7 ± 11.1 / 0.4 ± 8.8 | **6/6** |
| Flux-organized | 74.0 ± 3.4 | 13.7 ± 3.8 / 4.3 ± 8.1 | **6/6** |

Consistent with this concentration, replacing the |CFL| channel by its within-graph mean on the highest-|CFL| edge decile increases final rollout MSE by 101.0% and 128.2% for the geometry-organized and flux-organized models, respectively. The equal-sized random-edge intervention produces a substantially smaller effect. This targeted sensitivity shows that the learned models use the conditioning signal selectively on high-transport edges.

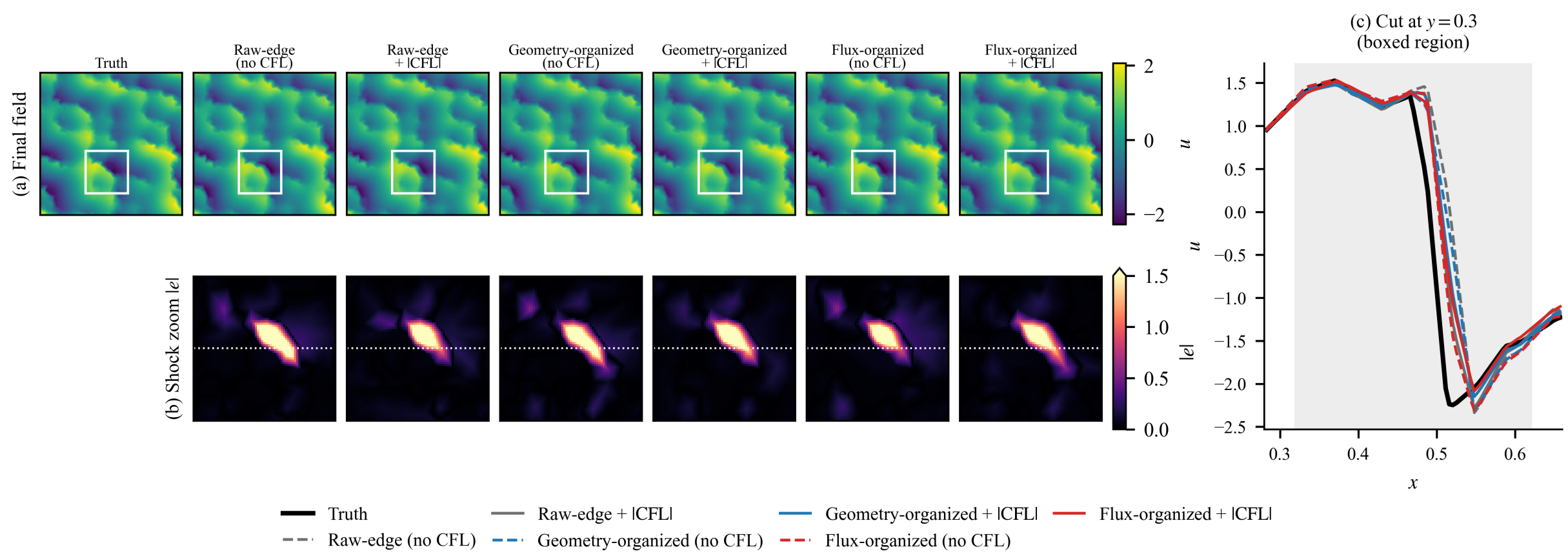

**Figure 4.** Representative larger-step OOD comparison on trajectory 041 (seed 42, KNN6, L4) after a 30-step autoregressive rollout for the six core models. (a) Final fields for the reference and the no-CFL/+ |CFL| pair in each feature family. (b) Absolute error in the fixed high-gradient box. (c) Cut through the box at y = 0.3. All predictions and the reference use the same snapshot index.

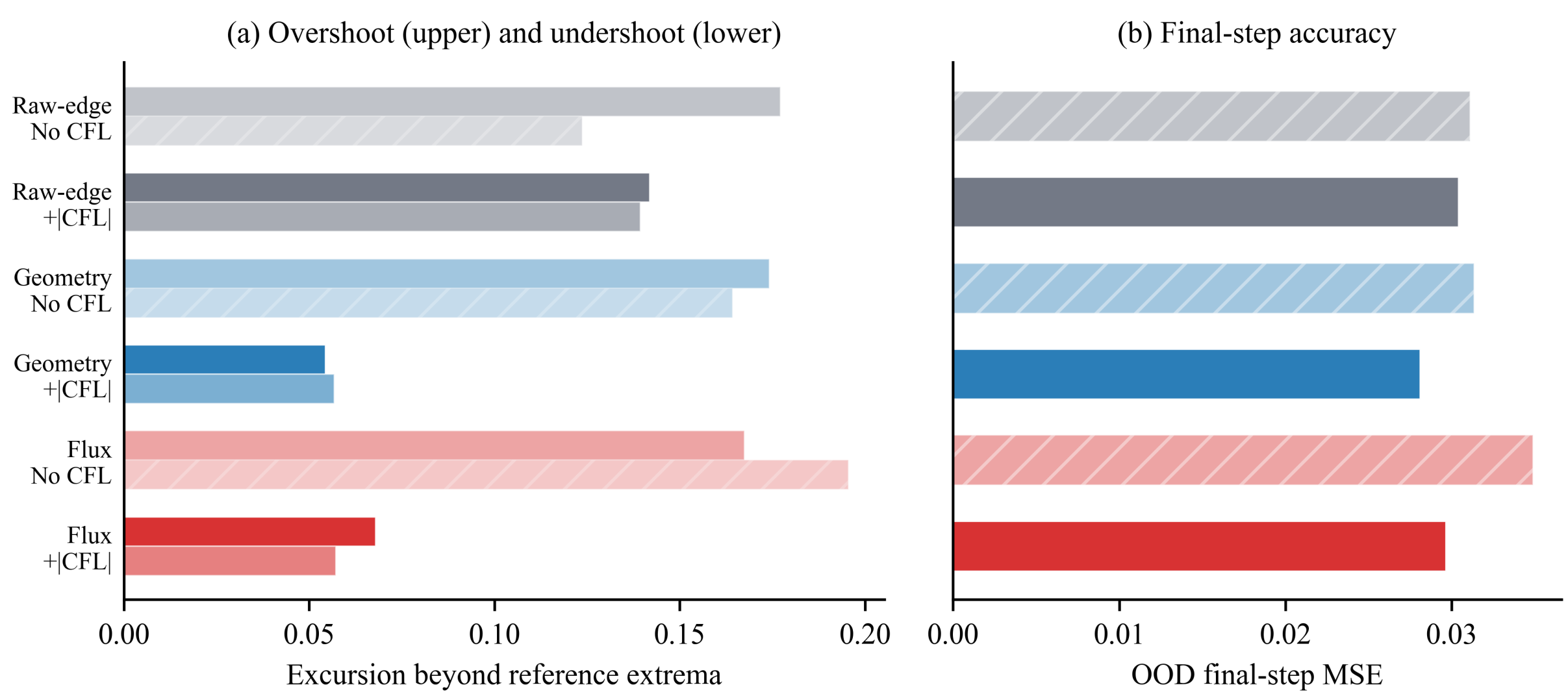

**Figure 5.** Shock-oscillation diagnostics and final-step accuracy for the six core models (larger-step OOD; L4, KNN4/6, seeds 42–44). (a) Final-step overshoot and undershoot beyond the reference extrema. (b) Final-step MSE. Within each family, the no-CFL parent and + |CFL| variant are shown as a matched pair.

## 5.4. Benefit grows with the transport gap and persists across message-passing depths

The final tests ask whether the gain from |CFL| conditioning grows as more transport must be represented in one learned update and remains distinct from merely enlarging the message-passing receptive field.

Table 11 extends the same trained models from snapshot CFL 0.85 to 1.0 and 1.2 using the same OOD initial conditions and reference solver; only the snapshot interval changes. Snapshot CFL 1.2 therefore probes a substantially larger transport gap within the same 30-step evaluation window.

The snapshot-CFL sweep shows distinct representation-dependent trends (Figure 6). The + |CFL| gain grows from 10.4% to 20.3% for geometry-organized and remains between 14% and 16% for flux-organized as snapshot CFL increases from 0.85 to 1.2. The raw-edge gain decays from 2.8% to 0.2%. Across all three snapshot CFL values, Geometry + |CFL| retains the lowest absolute final-step MSE among the transport-organized variants.

**Table 11.** Snapshot-CFL sweep for the six core models (L4, KNN4/6, seeds 42–44). The same trained models are evaluated at snapshot CFL 0.85, 1.0, and 1.2; only the snapshot interval changes.

| Model | Final MSE (0.85) | Final MSE (1.0) | Final MSE (1.2) | Rel gain (0.85/1.0/1.2) |
|---|---|---|---|---|
| Raw-edge | 0.0311±0.0072 | 0.0509±0.0098 | 0.0776±0.0100 | – |
| Raw + $\lvert CFL_{ij} \rvert$ | **0.0304±0.0080** | **0.0503±0.0113** | **0.0774±0.0120** | +2.8/+1.5/+0.2% |
| Geometry-organized | 0.0313±0.0068 | 0.0525±0.0078 | 0.0815±0.0074 | – |
| Geometry + $\lvert CFL_{ij} \rvert$ | **0.0281±0.0071** | **0.0449±0.0103** | **0.0651±0.0117** | +10.4/+14.9/+20.3% |
| Flux-organized | 0.0349±0.0076 | 0.0583±0.0097 | 0.0897±0.0095 | – |
| Flux + $\lvert CFL_{ij} \rvert$ | **0.0296±0.0054** | **0.0499±0.0075** | **0.0751±0.0075** | +14.4/+14.2/+16.1% |

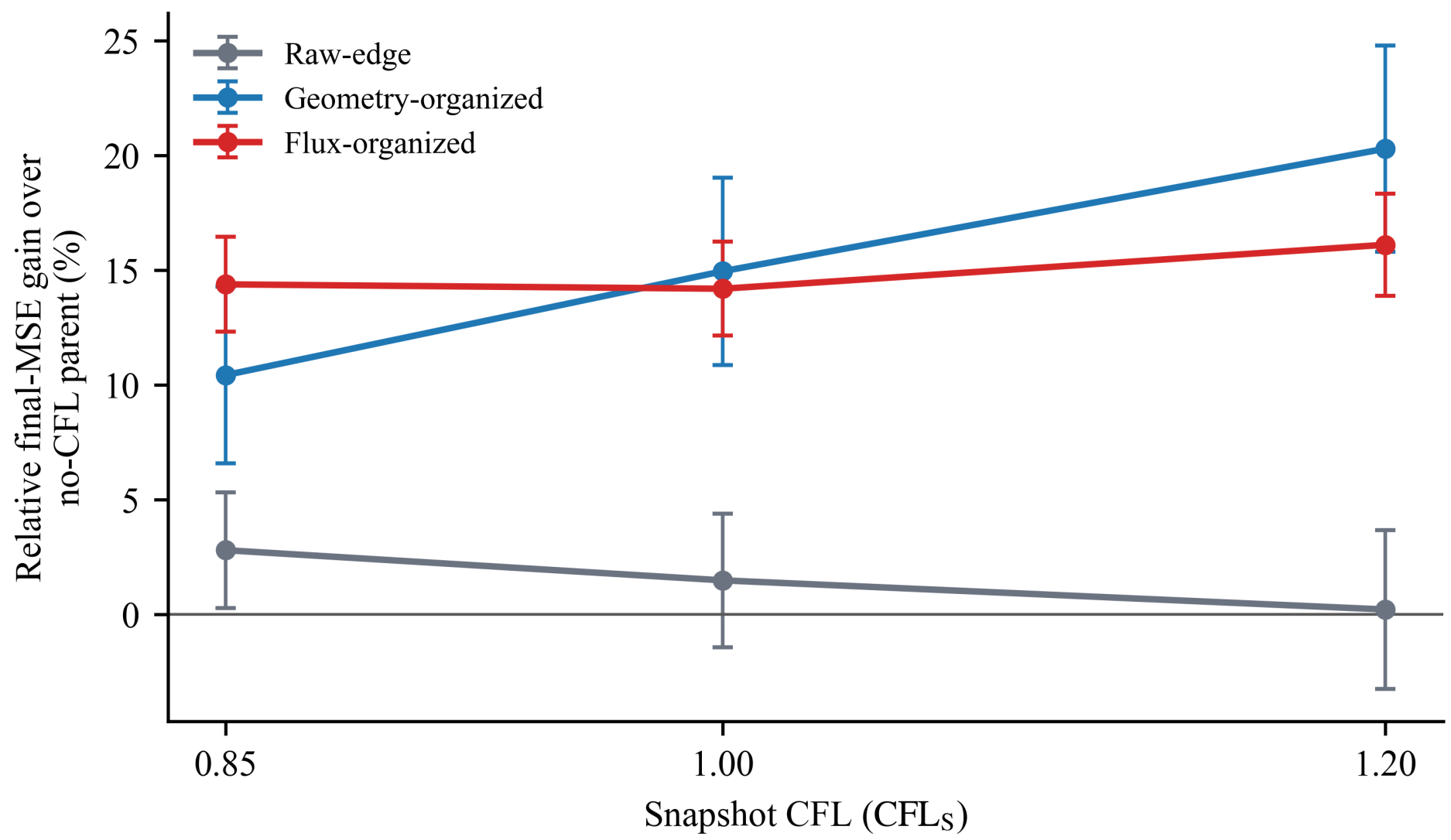

**Figure 6.** Within-family relative gain in final-step MSE from + |CFL| conditioning as a function of snapshot CFL (L4; mean with SEM over six matched settings). Geometry-organized and flux-organized gains persist or grow with the transport gap, whereas the raw-edge response approaches zero.

Finally, we vary message-passing depth L = 1, 2, 4, and 6 to ask whether a larger receptive field can replace explicit edge-local CFL conditioning (Figure 7).

The within-family |CFL| gain nevertheless persists at every evaluated depth in the transport-organized families. A larger receptive field therefore does not replace explicit edge-local CFL conditioning.

These trends appear across all three graph realizations and both connectivities and remain visible at the shallower depths, rather than depending on a single seed, connectivity, or receptive field.

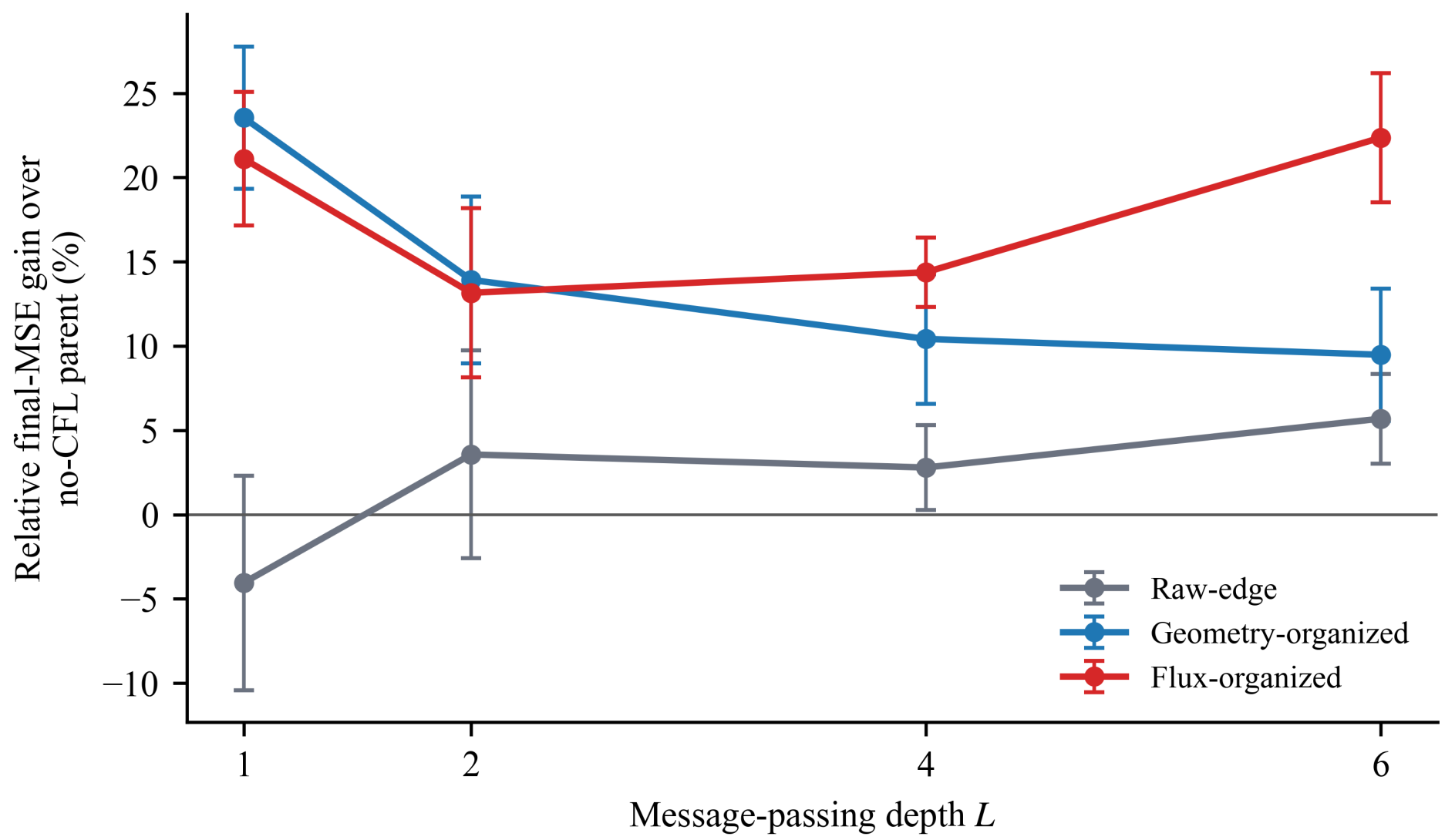

**Figure 7.** Within-family relative gain in final-step MSE from + |CFL| conditioning versus message-passing depth L for larger-step OOD rollouts (mean with SEM over six matched seed-KNN settings).

# 6. Discussion

## 6.1. Why geometry factorization makes CFL conditioning usable

The main bottleneck is not a lack of local transport information, but the effort required to assemble the available quantities into a transport update that the message function can use. The raw-edge inputs already contain endpoint states, edge displacement, edge length, the flux weight, and the snapshot interval. They are not, however, organized directly by interaction direction and transport scale.

The raw displacement already contains both edge direction and distance, but it couples them in a single vector. Collinear edge vectors can share an orientation while having different lengths, whereas orthogonal vectors can have the same length while representing different orientations. The message function must therefore recover unit direction and spatial scale before assigning them distinct roles in the update.

Hyperbolic transport is inherently directional. In classical shock-capturing schemes, upwinding and Riemann solvers use characteristic propagation to determine how information crosses an interface. The present representation does not impose an upwind flux or a Riemann solver. Instead, it exposes analogous local ingredients to message passing: the edge unit vector identifies the orientation of the directed interaction, the edge length sets its spatial scale, and the projected characteristic speed determines transport along that direction.

Direction and distance consequently enter the learned update in different ways. Direction determines how the characteristic velocity projects along the edge, whereas edge length sets the local distance against which one-step transport is measured. The geometry-organized representation performs this factorization before learning, and CFL magnitude supplies the dimensionless transport scale of the same directed interaction over one snapshot update. The message function can therefore use CFL directly as an edge-specific scaling signal without first reconstructing the interaction direction and spatial scale from the raw displacement.

This enabling role differs from absolute model quality. Geometry + |CFL| achieves the lowest final-step error and is the preferred representation for the present problem. The flux-organized family shows a larger within-family response, but its no-CFL parent is less accurate and its matched variant does not outperform Geometry + |CFL|. Flux organization therefore serves as a diagnostic of conditioning sensitivity rather than a more complete final model.

The correspondence controls support the same mechanism. Shuffling preserves the distribution of the CFL channel while breaking the association between each value and the directed edge on which it was computed. The resulting loss of conditioning benefit shows that the model uses CFL as the scale of a particular edge interaction rather than as a graph-wide scalar.

These results support the compatibility hypothesis: CFL conditioning is effective when the edge representation explicitly exposes the direction and spatial scale of the local interaction quantified by the CFL, and when each CFL value remains associated with the same directed edge interaction on which it was computed. The key design choice is to organize the conditioning variable and interaction representation around the same transport structure.

This mechanism predicts that the gain from CFL conditioning should concentrate where a learned update must represent large local transport. Section 6.2 examines this prediction in the shock region.

### 6.2. Selective oscillation control at high-transport shock interactions

Global MSE alone cannot distinguish accurate shock transport from a prediction that reduces error by broadly smoothing the front. The transport-stratified analysis in Section 5.3 (Table 10) isolates where, within the shock region, the conditioning gain actually occurs.

As reported in Table 10, error under the no-CFL parent is concentrated predominantly in the higher-transport half of shock-region nodes, and conditioning corrects that same subset disproportionately. This concentration is the expected signature of the compatibility mechanism: if |CFL| acts as an edge-specific update scale, its benefit should scale with how much transport that edge must represent, and the shock region is precisely where per-step transport is largest and errors compound fastest during rollout.

The evaluation-time ablation (Section 5.3) provides complementary evidence towards the same conclusion that degrading the conditioning signal specifically on the highest-transport edges is far more damaging than an equivalent random perturbation. This asymmetry indicates that the network has not learned to treat |CFL| as a generic auxiliary input it can ignore under distribution shift. Instead, it has learned to depend on the value at exactly the edges where the transport scale it encodes is the largest, which is the behavior the compatibility hypothesis predicts rather than a coincidental byproduct of adding an extra channel.

At these high-transport interactions, an error in the local update scale can accumulate into shock-position error and overshoot or undershoot during autoregressive rollout. Because geometry organization exposes the interaction direction and spatial scale while |CFL| supplies its update scale, the two together let the message function regulate exactly these interactions more consistently. Without this pairing, the same information is present but not addressable by edge. The observed gain is not a formal stability guarantee. In the geometry- and flux-organized families, the 10.4–14.4% reductions in global final-step MSE are accompanied by lower shock-region MSE and reduced overshoot and undershoot. These results link the overall error reduction to improved prediction quality near steep fronts.

### 6.3. Scope, limitations, and future directions

The main ablation uses the WENO5/SSP-RK3 anisotropic scalar Burgers benchmark to isolate the compatibility mechanism in a single-characteristic-family setting. Appendix A evaluates the same geometry-organized edge representation with edge-local CFL-magnitude conditioning on a shock-dominated two-dimensional shallow-water system with three coupled characteristic speeds. Geometry + |CFL| retains a clear advantage over its no-CFL parent and a matched MeshGraphNet-style baseline [10], showing that the observed benefit is not restricted to the scalar testbed.

The model predicts node increments on a periodic point-cloud graph rather than conservative fluxes on a control-volume mesh. The architecture does not enforce exact discrete conservation. The edge-local transport quantities used here are therefore conditioning features, not numerical fluxes that satisfy a discrete conservation law.

Appendix A.4 evaluates the trained Geometry models without retraining on unseen periodic point-cloud graphs with different node arrangements and node counts. The average CFL-conditioning benefit persists across these graph conditions. All experiments also use one message-passing architecture and the same reference-data pipeline.

The shock diagnostics and interventions identify where the model uses the conditioning signal and how the rollout changes. They do not establish formal stability, monotonicity, or conservation.

Future work should test this design principle in conservative graph updates, additional hyperbolic systems, and broader graph distributions.

## 7. Conclusions

This study examined when edge-local CFL conditioning becomes useful for GNN rollouts of shock-forming transport over larger snapshot intervals, and found that the answer depends on representation, not feature count (Sections 5.1–5.2). What matters instead is organizing the existing information into an explicit interaction direction, a spatial scale, and an update scale that the message function can act on directly. CFL conditioning is most effective when the edge representation already exposes the direction and spatial scale of the interaction it is meant to modulate, and when each CFL value stays attached to the specific edge it was computed for (Sections 5.2 & 5.3).

Among the representations tested, geometry organization paired with |CFL| gives the lowest absolute error, since it exposes unit direction and edge length separately while |CFL| supplies the dimensionless scale of that same directed interaction (Section 6.1). The flux-organized family shows a stronger within-family conditioning response, but does not improve on this best absolute accuracy, so it serves as a useful diagnostic of conditioning sensitivity rather than the preferred final representation.

The gain is not diffusive, as it concentrates at high-transport shock interactions, reduces oscillation rather than smearing the front, and persists as the receptive field grows or the transport gap widens (Sections 5.3 & 5.4), which can be viewed as evidence for a genuinely edge-local mechanism rather than a global regularization effect. The broader implication for GNN-based PDE solver design is that edge features should be organized around the physical interaction a conditioning variable is meant to modulate, rather than expanded in raw form and left for message passing to reassemble. This principle was tested primarily in a non-conservative architecture on a single-characteristic-family law with matched graph realizations and further examined on a coupled shallow-water system and unseen periodic graphs in Appendix A; extending it to conservative update rules, additional hyperbolic systems, and broader graph distributions (Section 6.3) is the natural next step for establishing how far the compatibility hypothesis generalizes.

# ACKNOWLEDGMENTS

This work was supported by the National Research Foundation of Korea (NRF) grant funded by the Korean government (Ministry of Science and ICT, MSIT) (No. RS-2024-00407902). This work was supported by the National Research Foundation of Korea (NRF) grant funded by the Korean government (MSIT) (No. RS-2023-NR077008). This work was also partially supported by the National Science Foundation under Grant No. DMREF-2203580.

# AUTHOR DECLARATIONS

## Conflict of Interest

J.B.C., H.S.U., and S.S.B. are co-founders of and hold equity in MaterIQ, Inc., which operates in the field of artificial intelligence and physics modeling. The research presented in this manuscript was conducted independently of MaterIQ, Inc., and the company had no role in the study design, data collection, or analysis. The remaining authors have no conflicts to disclose.

## Author Contributions

Jinyoung Jeong: Conceptualization; Methodology; Software; Validation; Formal analysis; Investigation; Data curation; Visualization; Writing – original draft.

Joseph B. Choi: Methodology; Writing – review & editing.

Xinlun Cheng: Methodology; Writing – review & editing.

H. S. Udaykumar: Formal analysis; Writing – review & editing; Supervision.

Sanghun Choi: Writing – review & editing; Supervision; Resources; Funding acquisition.

Stephen S. Baek: Conceptualization; Methodology; Formal analysis; Writing – review & editing; Supervision; Project administration; Resources.

# DATA AVAILABILITY

The datasets and source code supporting this study are publicly available at https://github.com/Jinyoung-Jeong/hyperbolic-gnn.

# APPENDIX A: ADDITIONAL VALIDATION

## A.1. Scalar control analyses

A smooth-regime nested-grid check of the scalar WENO5/SSP-RK3 implementation yielded observed convergence orders of 4.3–4.9. For each matched setting, the within-family |CFL| gain was defined as the parent-relative reduction in final-step MSE obtained by adding |CFL|. Relative to the Raw-edge gain, the Geometry- and Flux-organized gains were larger in all six settings, by mean differences of 7.6 and 11.6 percentage points, respectively (two-sided exact Wilcoxon signed-rank $p = 0.031$ for each comparison).

To distinguish CFL information from the addition of one extra input channel, Geometry + |CFL| was also compared with the dimension-matched random-noise control defined in Section 3.1. Geometry + |CFL| reduced final-step MSE by 5.9% on average relative to this control and achieved the lower error in all six matched settings (two-sided exact Wilcoxon signed-rank $p = 0.031$). The comparison is reported directly rather than subtracting the noise-control gain from the no-CFL comparison.

To assess sensitivity to optimization randomness, the Geometry no-CFL and Geometry + |CFL| comparison was repeated with three independent training seeds for each of the six graph settings. Geometry + |CFL| achieved lower final-step MSE in all 18 paired runs; after averaging the training repeats within each graph setting, the mean reduction was 12.9%, with improvements in all six settings (two-sided Wilcoxon signed-rank $p = 0.031$).

Because the scalar initial fields have zero continuum mean, relative drift normalized by the initial mean is ill-conditioned. We therefore measured the final graph-mean discrepancy relative to the sampled reference and normalized it by the initial graph RMS. Across the 48 Geometry + |CFL| OOD rollouts, this excess graph-mean error averaged 0.64%; the corresponding predicted and reference graph-mean drifts were 0.71% and 0.61%, respectively. These uniform-node graph means are empirical conservation proxies rather than control-volume integrals. For the quadratic scalar flux, the edge-normal speed in Eq. (8) equals the Rankine–Hugoniot secant speed between the two endpoint states, projected along the directed edge. The edge-local CFL therefore carries a physically relevant shock-transport scale, although the learned node update does not itself enforce the Rankine–Hugoniot jump condition or exact discrete conservation.

## A.2. Coupled shallow-water extension

We evaluated the geometry-organized edge representation with edge-local CFL-magnitude conditioning on the inviscid two-dimensional shallow-water system governed by

$$\frac{\partial \mathbf{q}}{\partial t} + \frac{\partial \mathbf{F}_x(q)}{\partial x} + \frac{\partial \mathbf{F}_y(q)}{\partial y} = 0 \tag{A1}$$

The conservative state and Cartesian fluxes are

$$\mathbf{q} = \begin{bmatrix} h \\ hu \\ hv \end{bmatrix}$$
$$\mathbf{F}_x(\mathbf{q}) = \begin{bmatrix} hu \\ hu^2 + \frac{1}{2}gh^2 \\ huv \end{bmatrix} \tag{A2}$$
$$\mathbf{F}_y(\mathbf{q}) = \begin{bmatrix} hv \\ huv \\ hv^2 + \frac{1}{2}gh^2 \end{bmatrix}$$

Here, $h$ is water depth, $u$ and $v$ are the depth-averaged Cartesian velocities, and $g = 1$. For an edge with unit edge direction $\mathbf{n} = (n_x, n_y)^{\mathrm{T}}$, the normal velocity is $u_n = un_x + vn_y$ and the three characteristic speeds are $u_n - c, u_n, u_n + c$, where $c = \sqrt{gh}$ [1,2]. The domain is the periodic unit square.

The edge quantities are evaluated from the arithmetic mean of the endpoint conservative states. The Geometry + |CFL| model supplies the magnitudes of the three characteristic families as a three-channel edge feature:

$$|\mathrm{CFL}_{ij}| = \frac{\Delta t}{d_{ij}}\left(|\bar{u}_{n,ij} - \bar{c}_{ij}|, |\bar{u}_{n,ij}|, |\bar{u}_{n,ij} + \bar{c}_{ij}|\right) \tag{A3}$$

The shock-dominated initial conditions set $u = v = 0$ over a background depth $h = 0.5$ and place three periodic circular depth humps at independently sampled locations on the unit square. Their radii are sampled uniformly from $[0.10, 0.20]$, their depth increments from $[1.0, 2.0]$, and each interface uses a smooth hyperbolic-tangent transition of width 0.015.

Reference trajectories are generated on a uniform 128 × 128 grid using componentwise WENO5 with global Lax–Friedrichs flux splitting, SSP-RK3 time integration, periodic boundary conditions, and a solver CFL of 0.25. The graph-learning protocol uses periodic jittered 32 × 32 point clouds (1024 nodes), with each coordinate perturbed by at most 0.35 lattice spacings, and symmetrized periodic KNN connectivity with K = 4 or 6. All compared GNN variants use hidden dimension 64, the same L4 processor and MLP architecture, and differ only in their edge-input construction. The protocol retains three graph seeds, a training snapshot CFL of 0.55, a 30-step rollout, and larger-step evaluation at snapshot CFL 0.85. The matched MeshGraphNet-style model uses the same training and evaluation protocol and serves as a backbone baseline rather than an official-code reproduction [10]. The MeshGraphNet-style baseline places the conserved state and the logarithm of the snapshot interval on nodes and uses only relative displacement and edge length on edges; unlike the Raw-edge model, it does not duplicate the endpoint states in the edge input.

A corresponding smooth-regime nested-grid check showed approximately fifth-order convergence; separate shallow-water validation cases preserved mass to machine precision and maintained positive water depth.

Table A1. Final-step errors for the shock-dominated two-dimensional shallow-water extension at snapshot CFL 0.85 (L4, KNN4/6, seeds 42–44; eight OOD trajectories per matched setting, 48 per model variant). Final-state and shock-region MSE are averaged over h, hu, and hv; depth overshoot and undershoot are the maximum positive and negative final-step excursions of h beyond the reference range, respectively.

| Model | Final-state MSE | Shock-region MSE | Water-depth overshoot | Water-depth undershoot |
|---|---|---|---|---|
| Raw-edge (No CFL) | 0.0311 | 0.0617 | 0.0305 | 0.00608 |
| Geometry-organized (No CFL) | 0.0158 | 0.0591 | 0.0275 | 0.00955 |
| Geometry + \|CFL\| | **0.00708** | **0.0352** | **0.00665** | 0.00815 |
| Matched MeshGraphNet-style | 0.0347 | 0.0523 | 0.0428 | **0.00330** |

Geometry + |CFL| reduces mean final-state MSE by 55.2% relative to its no-CFL parent and by 79.6% relative to the matched MeshGraphNet-style baseline. It also reduces shock-region MSE from 0.0591 to 0.0352 and depth overshoot from 0.0275 to 0.00665 relative to its no-CFL parent. All 48 Geometry + |CFL| rollouts remained finite and maintained positive water depth. The learned update is not conservative by construction; its mean graph-mass drift was 1.23%, so these measurements provide an empirical check rather than a formal conservation guarantee. Here, graph mass denotes the uniform-node mean of h because control-volume weights are unavailable on the irregular graph.

The excursion response is strongly asymmetric. From the means in Table A1, Geometry + |CFL| reduces water-depth overshoot by 75.8%, with lower overshoot in all six matched settings, whereas undershoot decreases by 14.6% and is lower in three settings. This asymmetry is consistent with the positive-hump initial conditions imposed over the background depth. A nodewise audit further shows that the highest-transport node decile, ranked by the mean incoming maximum wave-family |CFL|, contains 35.1% of the Geometry no-CFL parent's total positive excess above the reference maximum but only 0.3% of its negative excess below the reference minimum. Because the reported excursion diagnostics are maxima over nodes, their reduction should be interpreted primarily as suppression of extreme spikes. The concurrent reduction in shock-region MSE indicates that the improvement is not confined to a single extremum. This spatial concentration is consistent with CFL-magnitude conditioning acting most strongly at high-transport fronts.

Figure A1 provides a representative 30-step OOD comparison of the final water-depth field, absolute water-depth error, and steep-front cross-section.

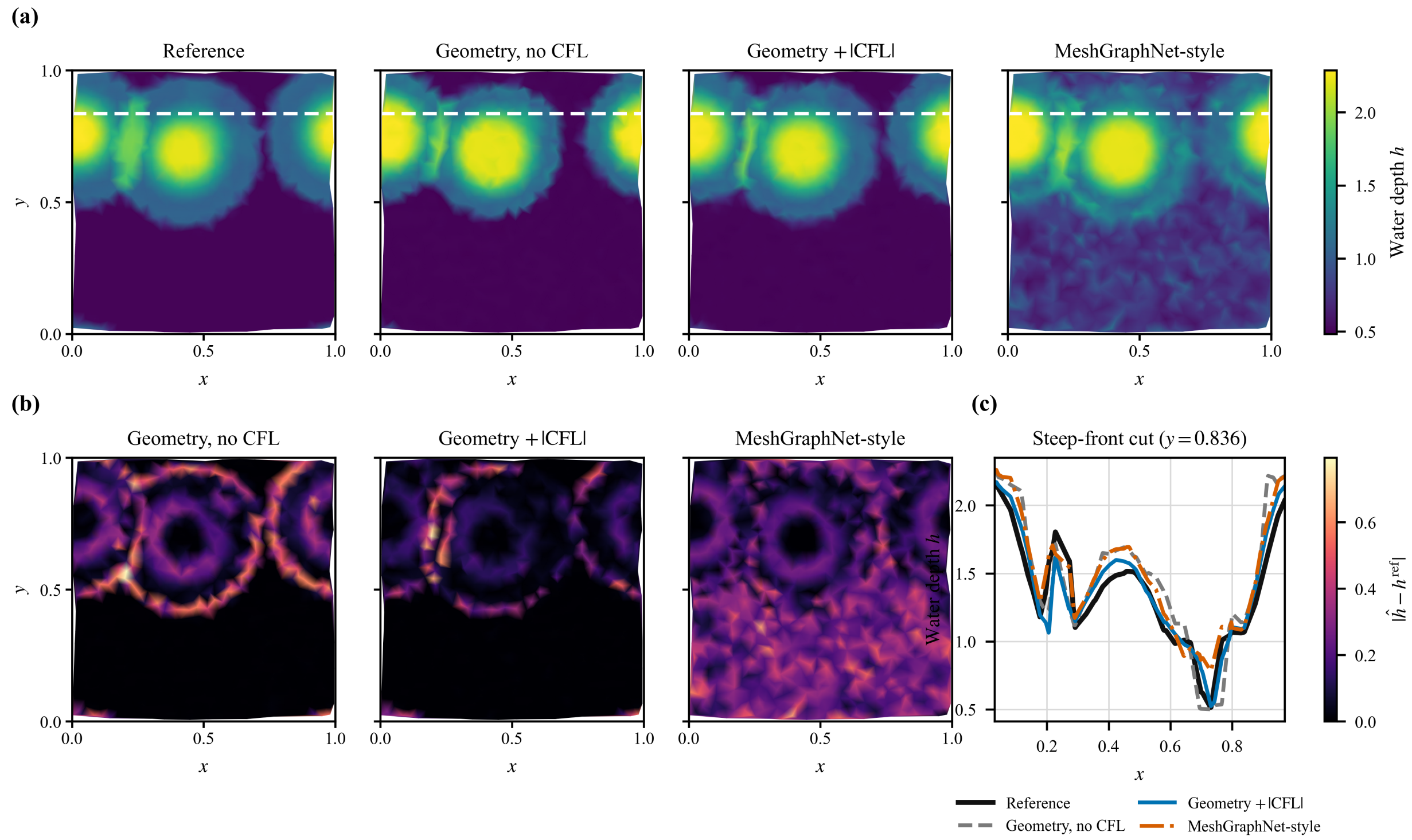

**Figure A1.** Representative 30-step OOD rollout for the shock-dominated two-dimensional shallow-water extension at snapshot CFL 0.85 (L4, seed 44, KNN6, trajectory 007). (a) Final water-depth fields h; the dashed line marks the cross-section at y = 0.836. (b) Corresponding absolute water-depth errors, $|\hat{h} - h|$, shown using a shared color scale. (c) Water-depth profiles along the marked cross-section. The displayed case has the parent-relative improvement closest to the median across the 48 OOD rollouts.

## A.3. Implementation-level wall-clock comparison

Using the Geometry + |CFL| model on a 1024-node KNN6 graph, an implementation-specific wall-clock benchmark compared one learned update on an NVIDIA RTX PRO 6000 Blackwell Max-Q GPU with one adaptive NumPy WENO5/SSP-RK3 advance on the server CPU over the same snapshot interval. After 10 warm-up runs, the median of 100 repetitions was 1.60 ms for the scalar GNN and 22.0 ms for WENO, corresponding to a 13.7-fold timing ratio; the corresponding shallow-water times were 1.33 ms and 71.0 ms, giving a 53.2-fold ratio. Because the methods use different hardware and implementations, these values are practical implementation-level comparisons rather than algorithm-independent speedups.

## A.4. Evaluation on unseen graphs

To test whether edge-local CFL conditioning remains useful beyond the training graph, we evaluated the original L4 Geometry no-CFL and Geometry + |CFL| checkpoints without retraining or fine-tuning. Each model was trained on 1024 nodes. Evaluation used the original graph and new periodic jittered point clouds with 576, 1024, and 2304 nodes, retaining KNN4/6 and the six matched settings. The eight OOD trajectories per setting, physical snapshot intervals, reference solver, and 30-step rollout were unchanged. The same reference trajectories were sampled on every target graph. The unseen 1024-node condition changes node placement alone; the other conditions also change graph spacing and the physical reach of the fixed-depth network.

Table A2. Cross-graph evaluation of the frozen Geometry models at snapshot CFL 0.85 (L4, KNN4/6, seeds 42–44; eight OOD trajectories per matched setting). Final-step MSE is averaged across the six settings. Mean paired gain is the average parent-relative MSE reduction within settings; Wins counts settings with lower MSE under |CFL| conditioning.

| Evaluation graph (nodes) | Geometry No CFL | Geometry + \|CFL\| | Mean paired gain (%) | Wins |
|---|---|---|---|---|
| Original (1024) | 0.03133 | **0.02808** | 10.4 | 5/6 |
| Unseen (576) | 0.03008 | **0.02646** | 12.1 | 5/6 |
| Unseen (1024) | 0.03768 | **0.03269** | 13.1 | 6/6 |
| Unseen (2304) | 0.07427 | **0.05842** | 21.1 | 6/6 |

Geometry + |CFL| retains lower mean final-step MSE than its no-CFL parent on all three unseen graph sizes (Table A2). Mean paired gains are 12.1%, 13.1%, and 21.1% for 576, 1024, and 2304 nodes, respectively. Absolute errors increase on the 2304-node graphs, while the conditioning advantage remains. These results extend the observed benefit to unseen node arrangements and graph sizes without retraining.

# REFERENCES

[1] R.J. LeVeque, Finite Volume Methods for Hyperbolic Problems, Cambridge University Press, 2002.

[2] E.F. Toro, Riemann Solvers and Numerical Methods for Fluid Dynamics, third ed., Springer, 2009.

[3] R. Courant, K. Friedrichs, H. Lewy, Über die partiellen Differenzengleichungen der mathematischen Physik, Mathematische Annalen 100 (1928) 32–74.

[4] S.K. Godunov, A difference method for numerical calculation of discontinuous solutions of the equations of hydrodynamics, Matematicheskii Sbornik 47 (89) (3) (1959) 271–306.

[5] P.L. Roe, Approximate Riemann solvers, parameter vectors, and difference schemes, Journal of Computational Physics 43 (1981) 357–372.

[6] A. Harten, P.D. Lax, B. van Leer, On upstream differencing and Godunov-type schemes for hyperbolic conservation laws, SIAM Review 25 (1983) 35–61.

[7] G.-S. Jiang, C.-W. Shu, Efficient implementation of weighted ENO schemes, Journal of Computational Physics 126 (1996) 202–228.

[8] C.-W. Shu, High order weighted essentially nonoscillatory schemes for convection dominated problems, SIAM Review 51 (2009) 82–126.

[9] A. Sanchez-Gonzalez, J. Godwin, T. Pfaff, R. Ying, J. Leskovec, P.W. Battaglia, Learning to simulate complex physics with graph networks, in: International Conference on Machine Learning, PMLR, 2020, pp. 8459–8468.

[10] T. Pfaff, M. Fortunato, A. Sanchez-Gonzalez, P.W. Battaglia, Learning mesh-based simulation with graph networks, in: International Conference on Learning Representations, 2021.

[11] J. Brandstetter, D.E. Worrall, M. Welling, Message passing neural PDE solvers, in: International Conference on Learning Representations, 2022.

[12] M. Lino, S. Fotiadis, A.A. Bharath, C.D. Cantwell, Multi-scale rotation-equivariant graph neural networks for unsteady Eulerian fluid dynamics, Physics of Fluids 34 (2022) 087110.

[13] Y. Bar-Sinai, S. Hoyer, J. Hickey, M.P. Brenner, Learning data-driven discretizations for partial differential equations, Proceedings of the National Academy of Sciences 116 (2019) 15344–15349.

[14] D. Kochkov, J.A. Smith, A. Alieva, Q. Wang, M.P. Brenner, S. Hoyer, Machine learning–accelerated computational fluid dynamics, Proceedings of the National Academy of Sciences 118 (2021) e2101784118.

[15] B. Stevens, T. Colonius, Enhancement of shock-capturing methods via machine learning, Theoretical and Computational Fluid Dynamics 34 (2020) 483–496.

[16] K. Um, R. Brand, Y. Fei, P. Holl, N. Thuerey, Solver-in-the-loop: Learning from differentiable physics to interact with iterative PDE-solvers, in: Advances in Neural Information Processing Systems, vol. 33, 2020, pp. 6111–6122.

[17] P.W. Battaglia, J.B. Hamrick, V. Bapst, A. Sanchez-Gonzalez, V. Zambaldi, M. Malinowski, A. Tacchetti, D. Raposo, A. Santoro, R. Faulkner, et al., Relational inductive biases, deep learning, and graph networks, arXiv preprint arXiv:1806.01261, 2018.

[18] K. Xu, W. Hu, J. Leskovec, S. Jegelka, How powerful are graph neural networks?, in: International Conference on Learning Representations, 2019.

[19] J. Gilmer, S.S. Schoenholz, P.F. Riley, O. Vinyals, G.E. Dahl, Neural message passing for quantum chemistry, in: Proceedings of the 34th International Conference on Machine Learning, PMLR 70, 2017, pp. 1263–1272.

[20] F. de Avila Belbute-Peres, T. Economon, Z. Kolter, Combining differentiable PDE solvers and graph neural networks for fluid flow prediction, in: Proceedings of the 37th International Conference on Machine Learning, PMLR 119, 2020, pp. 2402–2411.

[21] L. Tesán, M.M. Iparraguirre, D. González, P. Martins, E. Cueto, On the under-reaching phenomenon in message passing neural PDE solvers: Revisiting the CFL condition, Computer Methods in Applied Mechanics and Engineering 449 (2026) 118476.

[22] S. Gottlieb, C.-W. Shu, E. Tadmor, Strong stability-preserving high-order time discretization methods, SIAM Review 43 (2001) 89–112.

[23] V.V. Rusanov, The calculation of the interaction of non-stationary shock waves and obstacles, USSR Computational Mathematics and Mathematical Physics 1 (2) (1962) 304–320.

[24] M. Fortunato, T. Pfaff, P. Wirnsberger, A. Pritzel, P.W. Battaglia, MultiScale MeshGraphNets, arXiv preprint arXiv:2210.00612, 2022.

[25] M. Horie, N. Mitsume, Graph neural PDE solvers with conservation and similarity-equivariance, in: Proceedings of the 41st International Conference on Machine Learning, PMLR 235, 2024, pp. 18785–18814.

[26] I. Loshchilov, F. Hutter, Decoupled weight decay regularization, in: International Conference on Learning Representations, 2019.

[27] R. Pascanu, T. Mikolov, Y. Bengio, On the difficulty of training recurrent neural networks, in: Proceedings of the 30th International Conference on Machine Learning, PMLR 28 (3), 2013, pp. 1310–1318.

[28] M. Takamoto, T. Praditia, R. Leiteritz, D. MacKinlay, F. Alesiani, D. Pflüger, M. Niepert, PDEBench: An extensive benchmark for scientific machine learning, in: Advances in Neural Information Processing Systems 35, Datasets and Benchmarks Track, 2022, pp. 1596–1611.

[29] F. Koehler, S. Niedermayr, R. Westermann, N. Thuerey, APEBench: A benchmark for autoregressive neural emulators of PDEs, in: Advances in Neural Information Processing Systems 37, Datasets and Benchmarks Track, 2024, pp. 120252–120310, doi: 10.52202/079017-3822.

[30] G. Zhao, M. Sun, A. Memmolo, S. Pirozzoli, A general framework for the evaluation of shock-capturing schemes, Journal of Computational Physics 376 (2019) 924–936.

[31] R. Ida, Y. Tamaki, S. Kawai, Theoretical link in numerical shock thickness and shock-capturing dissipation, Journal of Computational Physics 505 (2024) 112901.

[32] D.I. Shuman, S.K. Narang, P. Frossard, A. Ortega, P. Vandergheynst, The emerging field of signal processing on graphs: Extending high-dimensional data analysis to networks and other irregular domains, IEEE Signal Processing Magazine 30 (3) (2013) 83–98.

[33] A. Sandryhaila, J.M.F. Moura, Discrete signal processing on graphs: Frequency analysis, IEEE Transactions on Signal Processing 62 (12) (2014) 3042–3054.